%% file: easychair.tex
\documentclass{easychair}

\usepackage{doc}

\usepackage{tikz}
\usepackage{subcaption}

\usepackage[numbers]{natbib}

\usepackage{amssymb}
\usepackage{todonotes}

\usepackage{xspace}

\newcommand{\dpll}{DPLL\xspace}
\newcommand{\dpllt}{{DPLL($T$)}\xspace}

\include{macros}

\title{Top-Down Knowledge Compilation for\\ Counting Modulo Theories}

\author{
    Vincent Derkinderen\inst{1,2}
\and
    Pedro Zuidberg Dos Martires\inst{3}
\and
    Samuel Kolb\inst{1,2}
\and
    Paolo Morettin\inst{4}
}

\institute{
  KU Leuven,
  Leuven, Belgium\\
  \email{$\{$vincent.derkinderen, samuel.kolb$\}$@kuleuven.be}
\and
   Leuven.AI,
   Leuven, Belgium\\
\and
   \"Orebro University,
   \"Orebro, Sweden\\
   \email{pedro.zuidberg-dos-martires@oru.se}\\
\and
   University of Trento,
   Trento, Italy\\
    \email{paolo.morettin@unitn.it}\\
 }

\authorrunning{Derkinderen, et al.}

\titlerunning{Top-Down Knowledge Compilation for Counting Modulo Theories}

\begin{document}

\maketitle

\begin{abstract}

    Propositional model counting (\#SAT) can be solved efficiently when the input formula is in deterministic decomposable negation normal form (d-DNNF).
    Translating an arbitrary formula into a representation that allows inference tasks, such as counting, to be performed efficiently, is called knowledge compilation.
    Top-down knowledge compilation is a state-of-the-art technique for solving \#SAT problems that leverages the traces of exhaustive \dpll search to obtain d-DNNF representations.
    While knowledge compilation is well studied for propositional approaches, knowledge compilation for the (quantifier free) counting modulo theory setting (\#SMT) has been studied to a much lesser degree.
    In this paper, we discuss compilation strategies for \#SMT.
    We specifically advocate for a top-down compiler based on the traces of exhaustive \dpllt search.
\end{abstract}

\section{Introduction}
\label{sec:introduction}

The key motivation of knowledge compilation research is to compile a logical formula into a target language whose properties allow for certain tasks to be performed in time polynomial in the representation size~\citep{darwiche2002map}.
For propositional model counting (\#SAT) problems, this representation is called deterministic decomposable negation normal form (d-DNNF)~\citep{darwiche2001dDNNF}.
\citeauthor{HuangD05}~realised that d-DNNF compilation can be achieved through exhaustive DPLL search by simply storing the search traces~\citep{Darwiche04,HuangD05}. Several state-of-the-art d-DNNF compilers are based on this principle, e.g. c2d~\citep{Darwiche04}, dSharp~\citep{MuiseMBH12Dsharp}, miniC2D~\citep{OztokD15miniC2D}, and D4~\cite{LagniezM17D4}.

The advantage of compiling to a d-DNNF formula instead of directly computing the (weighted) model count, arises with repeated inference. 
As the d-DNNF only depends on the logical formula, any weight change does not require recompiling. This splits model counting into a hard offline compilation step, the cost of which is amortized over (cheap) repeated online steps. This has for example been exploited in probabilistic logic programming for parameter learning and efficiently answering multiple queries~\citep{FierensBRSGTJR15,ManhaeveDKDR21}.

Top-down knowledge compilers, which are based on exhaustive \dpll, have so far mainly focused on \#SAT while ignoring the domain of modulo theories~\cite{barrett2018satisfiability}.
In this paper we consider counting modulo theories (\#SMT) and discuss compilation strategy choices for the quantifier free setting (Section~\ref{sec:mod_options}).
Most importantly, we draw parallels with \#SAT and identify a \dpllt \cite{Nieuwenhuis2006} based approach as a promising direction for a future knowledge compiler.

\section{Background}
\label{sec:background}

\paragraph{\dpll traces \& d-DNNF.} The \dpll algorithm, introduced to solve the satisfiability problem (SAT), repeatedly branches on variables in order to find a satisfying assignment to all propositional Boolean variables~\citep{DavisP60DPPL,DavisLL62DPPL}.  It is clear that any $\lor$-node following from a branching decision satisfies the \emph{determinism} property: the children of an $\lor$-node do not share any models. This is important for model counting and model enumeration as to avoid considering a model more than once. %
The exhaustive version of the DPLL algorithm (\#DPLL) is an extension that does not stop when a model is found, instead continuing to find all models. 
This version, which is used for model counting, enumeration and compilation problems, often includes \emph{component decompositioning}.
This algorithmic improvement detects when there are sets of clauses that do not share any variables, i.e., \emph{components}, and solves those independently~\citep{Bacchus2003}. 
The process results in an $\land$-node that satisfies the \emph{decomposability} property: the children of $\land$-nodes do not share any variables. 
This property is important for task decomposability, for example in weighted model counting it ensures that each literal weight is considered at most once. 
Decomposability also avoids having $\land$-branches that could disagree on literals, which is important to avoid enumerating or considering logically inconsistent assignments.
Additionally, the traces are also in negation normal form (NNF) because negation only occurs on literals and no other operations than $\lor$ and $\land$ take place. Thus, the trace of an exhaustive \dpll algorithm naturally yields a d-DNNF formula~\citep{Darwiche04,HuangD05}.

\paragraph{Counting Modulo Theories.} The satisfiability modulo theory (SMT) problem extends propositional satisfiability (SAT) with respect to a certain decidable background theory $T$~\citep{barrett2009handbook}. Notable examples of theories $T$ include linear real (\lra) or integer (\lia) arithmetic and fixed-size bit vectors (\bv). 
In SMT-\lra, formulas include propositional and \lra atoms of the form $\sum_i c_i x_i \bowtie c$ with $c_i,c \in \mathbb{Q}$, $x_i$ a real variable, and $\bowtie \ \in \{<,>,\leq,\geq,=,\neq\}$. An example \lra formula $F$ would be $[(x \leq 0) \lor (x \geq 1)] \land [A \lor (x \leq 0)]$.
The task is determining whether there exists an assignment to the variables such that both the input formula $F$ and the background theory $T$ are satisfied. For instance, in the example above, $(x \le 1)$ entails $A$.
Just as SMT generalizes SAT with additional theories, counting modulo theories (\#SMT) generalizes model counting (\#SAT) with additional theories.
In stark contrast with the purely propositional setting, formulas involving numerical theories may have infinitely many models.
While in this paper we mainly consider the quantifier-free \lra setting, we expect the proposed ideas to apply to a broader set of quantifier-free theories.

\paragraph{\dpllt} is a generalisation of the \dpll algorithm that is designed to solve SMT problems~\citep{Nieuwenhuis2006}. The key difference with the classic \dpll algorithm is that it involves a theory specific solver that interacts with the \dpll algorithm, using the background theory $T$ to propagate additional atoms when possible. For instance in the previous example, when a branch assigns $(x \geq 1)$ to true, the theory solver would additionally propagate $(x \leq 0)$ to be false. We note that most current implementations are actually extensions of the conflict driven clause learning (CDCL) algorithm, which augments the \dpll algorithm with conflict clause learning and backjumping~\citep{MouraB08Z3,BayardoS97CDCL,Nieuwenhuis2006}. For the purpose of our discussion we will use \dpllt to refer to both \dpllt and its CDCL($T$) augmentation.

\section{d-DNNF for modulo theory}
\label{sec:ddnnf}

We now discuss in more detail the d-DNNF properties~\citep{darwiche2002map} in the context of \#SMT.

A propositional formula is in \textbf{negation normal form} iff negation only occurs on literals, and the only other operations are $\lor$ and $\land$. 
Similar to the \dpll algorithm, the traces of a \dpllt algorithm produce NNF formulas. The fact that atoms can be associated with additional logic (e.g. $\lra$ atoms such as $x {<} 5$), and that a theory specific solver aids the propagation process, does not change this. %
The traces of a \dpllt algorithm are also \textbf{deterministic}, i.e., the children of each $\lor$-node do not share any models, because every decision associated with $\lor$ still partitions the models. 
The \textbf{decomposability} property is defined as $\land$-node branches not sharing any variables. In the modulo theory setting this needs further specification, because different atoms can still be linked together through their inner logic (e.g. shared continuous variables).

\paragraph{Decomposability} on the level of atoms is insufficient to ensure the decomposability of counting tasks on the formula. 
Consider the following example where the $\land$-branches do not share any atoms, but do share variables: $[(x < 5) \lor (x < y)] \land [(x > 10) \lor (y > 10)]$.
Here, $\{(x < 5),(x > y)\}$ and $\{(x > 10),(y > 10)\}$ are satisfying truth assignments for respectively the left and right $\land$-branch. Together however, they do not form a theory satisfying truth assignment. 
The benefit of decomposing on the level of variables, is that the previous situation is impossible, making it easier to ensure that each truth assignment is indeed theory consistent.
Unfortunately such decomposability can not be achieved in general. 
For the purpose of our discussion on modulo theory compilation, we will therefore use the d in d-DNNF to refer to decomposability on an atomic level rather than on a variable level. 
We will however expect that all truth assignments captured by the compiled formula are theory satisfiable. 
This is not necessarily ensured when using a theory agnostic compilation approach, which we discuss in more detail in the next section. For now, note that the traces of an exhaustive \dpllt algorithm will produce exactly such d-DNNF formulas where each truth assignment is theory satisfiable.

\section{Compilation strategies}
\label{sec:mod_options}

\subsection{Theory aware versus theory agnostic}\label{subsec:aware_agnostic}
Approaches for compiling a formula with respect to a background theory $T$ can be divided into two categories: theory aware versus theory agnostic approaches. In case of the latter, 
a Boolean abstraction of the input formula $F$ is first created, i.e. the formula that is obtained by replacing every theory atom in $F$ with a fresh Boolean atom. 
Afterwards, the Boolean abstraction can be compiled into a d-DNNF formula using any off-the-shelf propositional compiler. 
As a consequential benefit, advancements made to those compilers are automatically inherited.
Evidently, theory agnostic approaches also have a downside compared to approaches that are more theory aware. Namely, since the compiler is unaware of $T$, the resulting d-DNNF formula may contain truth assignments that are not consistent with $T$. For example, when abstracting $F = (x \leq 0) \lor (x \geq 1)$, both literals could be assigned to true, forming a $T$-unsatisfiable truth assignment. 
It is then up to the down stream inference algorithm that uses the resulting d-DNNF, to deal with the inconsistent models, typically impacting negatively the run time of the downstream task. 
For example, this theory agnostic approach was adopted by~\citeauthor{KolbMR19FXSDD} for solving weighted model integration (WMI) problems, i.e. weighted \#SMT problems over \lra formulas~\citep{KolbMR19FXSDD}. 
As a consequence of the theory-agnostic offline phase, a potentially large number of intermediate computations during the online integration procedure would result in regions with no models and would be discarded.

Figure~\ref{fig:smaller} shows an example of a formula representation that may result from a theory agnostic approach (\ref{fig:smaller1}), compared to a theory aware approach (\ref{fig:smaller2} and \ref{fig:smaller3}). In this example, the formula representation shrank once the theory was considered. This is not always necessarily the case: the effect that theory awareness has on the representation size heavily depends on the exploitable structure, which may increase or decrease. Depending on the application requirements, we could even consider a more condensed representation that omits theory implied atoms (Figure~\ref{fig:smaller3}).

\tikzstyle{union} = [rounded corners,text centered, draw=black]
\tikzstyle{intersection} = [rounded corners,text centered, draw=black]
\tikzstyle{times} = [rounded corners,text centered, draw=black]
\tikzstyle{plus} = [rounded corners,text centered, draw=black]
\tikzstyle{int} = [circle, text centered, draw=black]
\tikzstyle{leaf} = [rounded corners, text centered, draw=black]
\tikzstyle{arrow} = [thick,->,>=stealth]

\begin{figure}
\centering
\begin{subfigure}{.4\textwidth}
    \centering
    \begin{tikzpicture}[node distance=.75cm]
        \node (plus1) [plus] {$+$};     %
        \node (times1) [times, below of=plus1, left of=plus1] {$\times$};   %
            \node (l1) [leaf, below of=times1, left of=times1] {$B_2$};
            \node (plus2) [plus, below of=times1, right of=times1, xshift=-.45cm] {$+$};
                \node (l2) [leaf, below of=plus2, left of=plus2] {$\neg B_1$};  %
                \node (times3) [times, below of=plus2, right of=plus2] {$\times$};   %
                    \node (l3) [leaf, below of=times3, left of=times3] {$B_1$};     %
                    \node (l4) [leaf, below of=times3, right of=times3] {$A$};      %
        \node (times2) [times, below of=plus1, right of=plus1] {$\times$};  %
            \node (l5) [leaf, below of=times2, right of=times2] {$\neg B_2$};

        \draw [arrow] (plus1) -- (times1);
        \draw [arrow] (plus1) -- (times2);
        \draw [arrow] (times1) -- (l1);
        \draw [arrow] (times1) -- (plus2);
        \draw [arrow] (plus2) -- (l2);
        \draw [arrow] (plus2) -- (times3);
        \draw [arrow] (times3) -- (l3);
        \draw [arrow] (times3) -- (l4);
        \draw [arrow] (times2) -- (l5);
        \draw [arrow] (times2) -- (times3);
    \end{tikzpicture}
    \caption{}
    \label{fig:smaller1}
\end{subfigure}%
\begin{subfigure}{.4\textwidth}
    \centering
    \begin{tikzpicture}[node distance=.75cm]
        \node (plus1) [plus] {$+$};
        \node (times1) [times, below of=plus1, left of=plus1] {$\times$};
            \node (l1) [leaf, below of=times1, left of=times1] {$B_2$};
            \node (l2) [leaf, below of=times1, right of=times1, xshift=-.45cm] {$\neg B_1$};
        \node (times2) [times, below of=plus1, right of=plus1] {$\times$};
            \node (l5) [leaf, below of=times2, right of=times2] {$\neg B_2$};
            \node (times4) [times, below of=times2, left of=times2, xshift=+.45cm] {$\times$};
                \node (l6) [leaf, below of=times4, left of=times4] {$B_1$};
                \node (l7) [leaf, below of=times4, right of=times4] {$A$};

        \draw [arrow] (plus1) -- (times1);
        \draw [arrow] (plus1) -- (times2);
        \draw [arrow] (times1) -- (l1);
        \draw [arrow] (times1) -- (l2);
        \draw [arrow] (times2) -- (l5);
        \draw [arrow] (times2) -- (times4);
        \draw [arrow] (times4) -- (l6);
        \draw [arrow] (times4) -- (l7);
    \end{tikzpicture}
\caption{}
\label{fig:smaller2}
\end{subfigure}%
\begin{subfigure}{.2\textwidth}
    \centering
    \begin{tikzpicture}[node distance=.75cm]
        \node (plus1) [plus] {$+$};
        \node (l1) [leaf, below of=plus1, left of=plus1] {$B_2$};
        \node (times2) [times, below of=plus1, right of=plus1] {$\times$};
            \node (l5) [leaf, below of=times2, left of=times2, xshift=+.45cm] {$B_1$};
            \node (l6) [leaf, below of=times2, right of=times2] {$A$};

        \draw [arrow] (plus1) -- (l1);
        \draw [arrow] (plus1) -- (times2);
        \draw [arrow] (times2) -- (l5);
        \draw [arrow] (times2) -- (l6);
    \end{tikzpicture}
\caption{}
\label{fig:smaller3}
\end{subfigure}
    \caption{Different representations of $(B_1 \lor B_2) \land (\neg B_1 \lor A)$, with abbreviations $B_1 = (x < y - 1)$, $B_2=(x > y + 1)$, and $A=(x>20)$.}
\label{fig:smaller}
\end{figure}
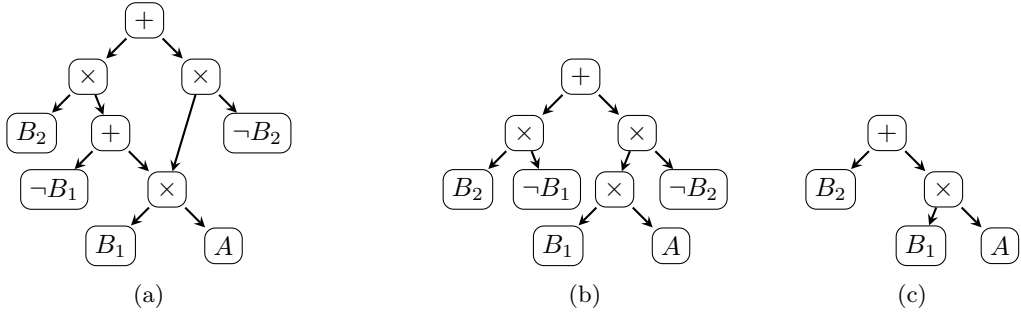

\subsection{Eager versus lazy solving}\label{subsec:lazy_eager}

It is possible to exploit the existing purely propositional tools while still achieving theory awareness, by first adapting the Boolean abstracted input formula $F$ in a way that no theory unsatisfiable truth assignments arise. This idea was used by early SMT solvers, where it is called `eager solving'~\citep{Nieuwenhuis2006}. For example, if $F$ includes both $(x \leq 0)$ and $(x \geq 1)$, which were respectively abstracted into $A_0$ and $A_1$, then also add $A_0 {\implies} \neg A_1$ or $A_1 {\implies} \neg A_0$ to $F$. Afterwards, because $F$ is purely propositional, any SAT solver (or in our case compiler) can be used. Since the adapted abstraction step is also present in eager SMT solvers, we can simply implement an eager theory compiler by using an existing eager SMT solver as foundation, swapping their SAT solver for a propositional d-DNNF compiler.

A major disadvantage of the eager approach is that depending on the theory $T$, it may introduce many additional variables, and greatly increase the formula size of the abstracted input formula $F$. For this reason, numerical theory SMT solvers have instead progressed towards using a so called `lazy solving'~\citep{Nieuwenhuis2006} approach that we discuss next. This does not mean, however, that the eager approach has become irrelevant. It still forms the state-of-the-art SMT approach for certain theories such as \bv\footnote{The 2022 SMT competition winner of the incremental quantifier free \bv track was Yices 2~\citep{Dutertre14Yices}, which uses bit blasting for \bv, an eager solving approach.}.
We hypothesize that similar conclusions can be drawn for the compilation setting, i.e., that a lazy approach is most suitable for numeric background theories.

In the most basic form, the lazy approach involves purely propositional reasoning over $F$, and verifying (partial) assignments with a theory solver to make sure that every model is indeed theory satisfiable. Enhancements of this approach have moved to a more proactive design where the theory solver instead helps propagating atoms and learning conflict clauses. This approach is the \dpllt (or CDCL($T$)) algorithm discussed in Section~\ref{sec:background} and is used by most state-of-the-art SMT solvers~\citep{BarbosaBBKLMMMN22Cvc5,ChristHN12,CimattiGSS13MathSAT,MouraB08Z3,Dutertre14Yices,HyvarinenMAS16OpenSMT}.

\subsection{Top-down versus bottom-up compilation}\label{subsec:top_bottom}

\dpll-based compilers are so called top-down compilers: they start the compilation process from the whole input formula and work their way down (by branching on literals). In contrast, bottom-up compilers process an input formula $F$ from the bottom of the expression to the top. For example, if $F$ is $(A \lor \neg B) \land (\neg A \lor C)$, then the two conjuncts are separately compiled before processing the conjunction itself. This design implies an apply-operation that performs conjunction, disjunction and negation. The apply-operation enables the incremental construction of (multiple) formulas, while a top-down compiler requires the complete formula to be known up front. To achieve an efficient apply-operation, however, the target language of bottom-up compilers is usually restricted to strongly deterministic and structured decomposable NNF, such as OBDD and the more general SDD class~\citep{darwiche2011sdd,pipatsrisawat2008new}, strict subsets of d-DNNF. In comparison, \dpll based top-down compilers instead produce Decision-DNNF, a strict superset of OBDD~\citep{DarwicheH20} (but not of SDD with which it partially overlaps).

An advantage of d-DNNF bottom-up compilation is that it does not restrict the input form. In contrast, top-down compilation is commonly restricted to formulas in conjunctive normal form (CNF). While it is possible to translate any propositional formula into CNF, this costs time and may result in an exponential expression (when using De Morgan's law), or requires introducing additional variables~\citep{tseitin1983complexity}.

An advantage of top-down compilation is that it has available the full information contained within $F$.
This assumption allows more informed decisions, in turn leading to more compact compiled formulas.
Oppositely, bottom-up compilation may have a pre-analysis step that considers the whole formula $F$ to determine a suitable guiding structure (cf. static variable ordering heuristics for OBDD~\citep{rice2008survey} or vtree heuristics for SDD~\citep{darwiche2011sdd}), but that structure can be suboptimal for intermediate formula representations. 
This naturally leads to the second point, which is that even when the final d-DNNF produced by bottom-up compilation is small, the intermediate results of the apply-operation may grow to be very large, severely impacting the run time and memory requirements~\citep{HuangD04}. 

In terms of compilation in the modulo theory domain, only bottom-up approaches have been considered so far. For example, the work on extended algebraic decision diagrams (XADD)~\citep{SannerDB11XADD} which uses a bottom-up approach. Within their implementation, they used a feasibility checker of a linear programming solver to prune theory unsatisfiable paths during- or after bottom-up compilation, observing impressive XADD size reductions. Other bottom-up compilers include work on difference decision diagrams~\citep{MollerLAH99DDD} and linear decision diagrams~\citep{ChakiGS09LDD}.

\section{Traces of an exhaustive \dpllt algorithm} 
\label{sec:proposed_solver}

In line with top-down compilation for the propositional setting, we propose to compile modulo theory formulas by storing the search traces generated by an exhaustive \dpllt algorithm. This is a theory aware, top-down, lazy approach, which we hypothesize to be very suitable for compiling formulas with numerical background theories such as \lra (cf. Section~\ref{subsec:lazy_eager}).

Existing SMT solvers based on the \dpllt framework form a natural implementation starting point. 
As explained in Section~\ref{sec:background}, the traces of such an approach form d-DNNF formulas.
To further augment the implementation, it is worth investigating optimisations developed for \#\dpll. For example, component decompositioning mentioned in Section~\ref{sec:background} is rarely used for satisfiability problems but has shown to be highly beneficial for model counting and compiling, especially in combination with caching~\citep{Bacchus2003}.
The augmentations are not straightforward however, because of the implicit interactions between atoms: 
components must be decomposed on a variable level (rather than atom), and previous decisions must be properly considered. 
Especially the latter is a novel challenge for caching.
Consider the following example formula $F'$: $\big(   (x < 3) \lor (x > 5)   \big) \land \big(   (y < 0) \lor (y > 4)   \big)$. Generally, branching on $(x > 5)$ does not imply $(y < 0)$. However, if $(x + y) < 5$ is a decision that has lead to the intermediate formula $F'$, then branching on $(x > 5)$ while compiling $F'$ does imply $(y < 0)$. This shows that the compiled form of $F'$ is indeed influenced by previous decisions.
This is in contrast to the propositional setting, where after making a decision $l$ and adapting formula $F$, the remaining formula $F|_{l}$ (here $F'$) is independent of previous decisions.

\paragraph{Related work}
To the best of our knowledge, we are the first to propose a theory aware top-down lazy compiler. The closest related works are \cite{braz2016probabilistic,MaLZ09}, who use a \dpllt like approach to count over modulo theories, but do not consider storing the traces for compilation. \cite{FeldsteinB21}, in contrast, does consider compilation but is positioned further from \dpllt, excluding optimizations such as conflict clause learning. Moreover, there is limited background theory support, for example atoms with arithmetic expressions such as $x + y < 2$ are not possible.

\section{Conclusion \& Future Work}
\label{sec:conclusion}

We have provided a discussion on compilation strategies for (top-down) d-DNNF compilation for quantifier free modulo theories.
We specifically propose compilation through an exhaustive version of the \dpllt algorithm, drawing parallels with the propositional setting's evolution from SAT to \#SAT. 
In future work, we aim to develop a tool based on these ideas, and plan to further investigate component decompositioning and caching for modulo theories. Additionally, refining the d-DNNF properties based on application requirements, is of interest.

\section*{Acknowledgments}
  This work was partially supported by the Research Foundation-Flanders (FWO, 1SA5520N and S007318N), the Flemish Government under the ``Onderzoeksprogramma Artifici\"ele Intelligentie (AI) Vlaanderen'' programme, the EU H2020 ICT48 project “TAILOR” under contract \#952215, and the Wallenberg AI, Autonomous Systems and Software Program (WASP) funded by the Knut and Alice Wallenberg Foundation. 
  PM is supported  by the PNRR project FAIR - Future AI Research (PE00000013),  under the NRRP MUR program funded by the NextGenerationEU. SK is supported by the ``Agentschap Innoveren \& Ondernemen'' (VLAIO) as part of the innovation mandate HBC.2021.0246.

\bibliographystyle{plainnat}
\bibliography{references}



\end{document}

%% file: macros.tex
\usepackage{xspace}

\newcommand{\bv}{\ensuremath{\mathcal{BV}}\xspace}
\newcommand{\lia}{\ensuremath{\mathcal{LIA}}\xspace}
\newcommand{\lra}{\ensuremath{\mathcal{LRA}}\xspace}